\documentclass{article}
\usepackage[preprint]{neurips_2025}
\usepackage{newtxtext}
\usepackage[colorlinks]{hyperref}
\usepackage{amsmath, amssymb}
\usepackage{tikz}
\usepackage{booktabs}
\usepackage{multirow}
\usepackage{graphicx}
\usepackage{listings}
\usepackage{xcolor}
\usepackage{algorithm}
\usepackage{algorithmicx}
\usepackage{algpseudocode}
\usetikzlibrary{arrows.meta, positioning, shadows, automata, quotes, backgrounds, fit, shapes, calc}
\lstdefinelanguage{MyDiff}{
  morecomment=[f][\color{red}]{-},   
  morecomment=[f][\color{green!50!black}]{+}, 
  morecomment=[f][\color{blue}]{@@}, 
  morecomment=[f][\color{gray}]{---}, 
  morecomment=[f][\color{gray}]{+++}  
}
\lstdefinelanguage{prompts}{
  moredelim=**[is][\color{blue}\bfseries]{@}{@}, 
}

\lstnewenvironment{promptlisting}[1][]{
  \lstset{
    language=prompts,
    basicstyle=\ttfamily\small,
    breaklines=true,
    breakatwhitespace=false,
    breakindent=1em,
    frame=single,
    tabsize=2,
    showstringspaces=false,
    captionpos=b,
    xleftmargin=0pt,
    xrightmargin=0pt,
    linewidth=\linewidth, 
    postbreak=,
    #1
  }
}{}

\newcommand{\method}[0]{$\textsc{COSMIR}$}
\newcommand{\methodfull}[0]{\textbf{C}hain \textbf{O}rchestrated \textbf{S}tructured \textbf{M}emory for \textbf{I}terative \textbf{R}easoning}
\title{COSMIR: Chain Orchestrated Structured Memory for Iterative Reasoning over Long Context}

\author{
\makebox[\textwidth][c]{%
    \begin{tabular}{c}
        Naman Gupta$^{1}$\thanks{Equal contribution.} \quad
        Shreeyash Gowaiker$^{2}$\footnotemark[1] \quad
        Arun Iyer$^{1}$ \quad
        Kirankumar Shiragur$^{1}$ \quad
        Ramakrishna B. Bairi$^{1}$ \\
        Rishikesh Maurya$^{1}$ \quad
        Ritabrata Maiti$^{1}$ \quad
        Sankarshan Damle$^{1}$ \quad
        Shachee Kumar Mishra$^{1}$ \\
        \\
        \textnormal{$^{1}$Microsoft} \\
        \textnormal{$^{2}$Work done during internship at Microsoft}
    \end{tabular}%
}
}

\begin{document}
\maketitle




\begin{abstract}
Reasoning over very long inputs remains difficult for large language models (LLMs). Common workarounds either shrink the input via retrieval (risking missed evidence), enlarge the context window (straining selectivity), or stage multiple agents to read in pieces. In staged pipelines (e.g., Chain of Agents, CoA), free-form summaries passed between agents can discard crucial details and amplify early mistakes. We introduce \method~(\methodfull), a chain-style framework that replaces ad hoc messages with a structured memory. A \textsc{Planner} agent first turns a user query into concrete, checkable sub-questions. \textsc{worker} agents process chunks via a fixed micro-cycle: Extract, Infer, Refine, writing all updates to the shared memory. A \textsc{Manager} agent then \textsc{Synthesizes} the final answer directly from the memory. This preserves stepwise read-then-reason benefits while changing both the communication medium (structured memory) and the worker procedure (fixed micro-cycle), yielding higher faithfulness, better long-range aggregation, and auditability. On long-context QA from the HELMET suite, \method~reduces propagation-stage information loss and improves accuracy over a CoA baseline.
\end{abstract}

\section{Introduction}

Large Language Models (LLMs) have rapidly advanced language understanding and generation tasks, supporting assistants, search, and retrieval systems~\citep{wu2024autogen,langchain}. However, reasoning over \emph{long} inputs, for example, books, extended technical documents, or large code repositories, remains brittle \citep{liu2024lost, brown20fewshot, srivastava2023imitationgamequantifyingextrapolating}. Mitigation strategies typically follow two paths. The first contracts the input through retrieval (e.g., RAG~\citep{lewis2020retrieval}), which can omit crucial evidence and inject noise. The second expands the model context windows \citep{peng2024yarn}, but still struggles with selectivity~\citep{liu2024lost} and faces practical scaling limits~\citep{wang2024beyond}.

A complementary line decomposes long-context reasoning into steps executed over shorter spans. This includes tree- or graph-structured prompting \citep{yao2023tree} and multi-agent coordination \citep{zhang2024chain}. Although effective, sequential pipelines that pass \emph{free-form summaries} between steps are vulnerable to compression loss and cascading errors. An agent must spot what matters in its local fragment, compress it into an ad hoc message, and anticipate future relevance. Omissions or imprecisions early on can silently propagate and degrade the final answer (Appendix~\ref{sec:coa_failure_modes}).

We propose \method, \textit{\methodfull}, a training-free framework that keeps the stepwise ``read–reason'' benefit while replacing free text messages with a \emph{structured, centralized working memory}. A \textsc{Planner} converts the user query into concrete, checkable sub-questions. \textsc{Workers} traverse chunks using a fixed micro-cycle: \textsc{Extract} evidence under a memory budget, \textsc{Infer} grounded claims from accumulated evidence, and \textsc{Refine} the unresolved question set. The worker then writes the information into a shared memory $M$. A \textsc{Manager} then \textsc{Synthesizes} the final answer directly from \(M\). This design reduces propagation stage information loss, improves long-range aggregation, and yields an auditable trace of how the answer was produced.

Our key contributions are: 1] We introduce \method, a training-free, interpretable framework for long-context reasoning that replaces free-form message passing with a centralized memory and a fixed worker micro-cycle. 2] We show in long-context QA benchmarks (HELMET suite~\citep{yen2025helmet}) that \method~reduces information loss and improves accuracy over a Chain of Agents (CoA)~\citep{zhang2024chain} baseline at comparable cost.

\paragraph{Paper organization.}
Section~\ref{sec:case-study} analyzes a representative failure of CoA~\citep{zhang2024chain} due to propagation stage information loss and illustrates how \method~prevents it. Section~\ref{sec:related-work} situates our work among long-context modeling, multi-agent prompting, and structured memory. Section~\ref{sec:method} formalizes \method~end-to-end describing both the structured, centralized memory and the different agent executions. Sections~\ref{sec:experimentsetup} and~\ref{sec:results} describe the experimental setup and results. We discuss limitations and future work before concluding.

\section{Example Case Study}
\label{sec:case-study}
\begin{figure}[ht]
    \centering    \includegraphics[width=\linewidth]{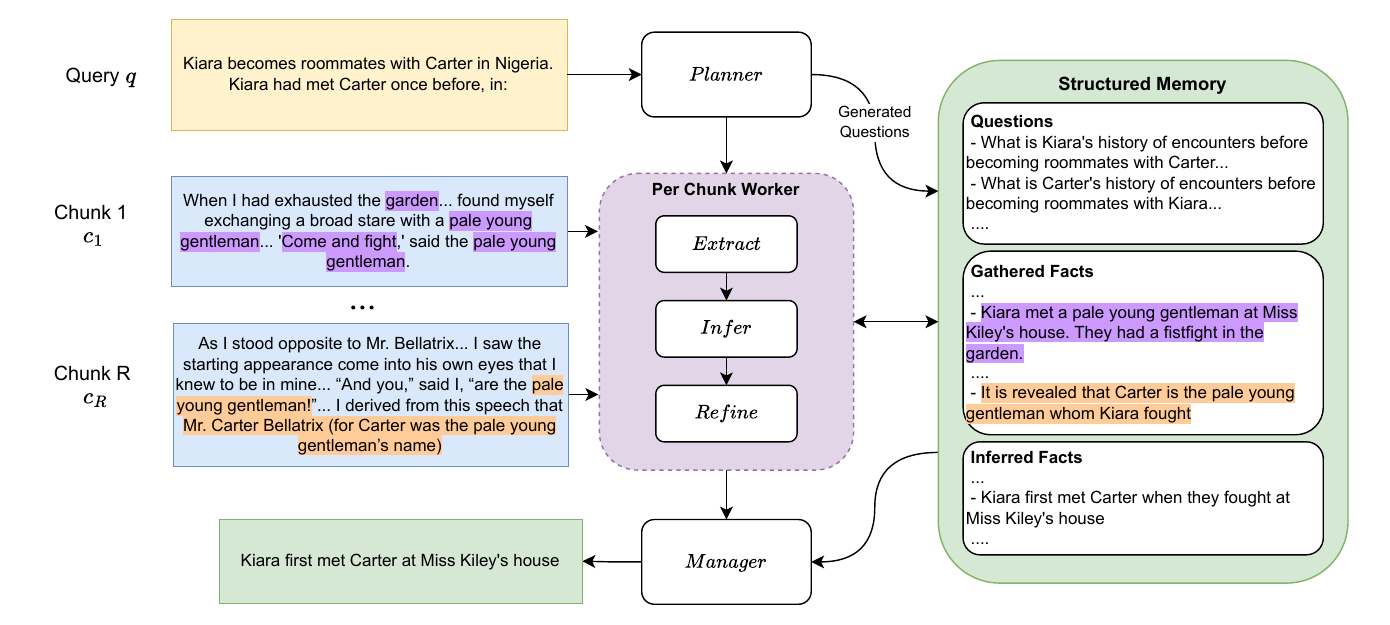}
    \caption{Overview of \method, a training-free framework for long context tasks. It consists of a \textsc{Planner} agent which given the question generates clarifying sub-questions. Segmented chunks from the context are then processed by the \textsc{Worker} agent in a fixed micro cycle which has three phases: \textsc{Extract}, \textsc{Infer} and \textsc{Refine}. Through these phases the \textsc{Worker} agent edits a structured centralized memory by extracting facts, making logical inferences over the facts and planning next steps by removing/adding new sub-questions. Finally, the structured memory is passed to the \textsc{Manager} agent to generate a final coherent answer. Boxes in blue are excerpts from chunks $c_1$ and $c_R$. Key portions of these excerpts, that are needed to answer the query $q$ have been highlighted and corresponding facts that have been extracted from these chunks have been highlighted in the structured memory.}
    \label{fig:coma-diag}
\end{figure}

Figure~\ref{fig:coma-diag} illustrates \method~on a question from the \textbf{InfBench-QA} dataset (Section~\ref{sec:experimentsetup}): \emph{Where did Kiara and Carter first meet before becoming roommates in Nigeria?} Early in the book (chunk~1), the text states that \textbf{Kiara met a pale young gentleman at Miss Kiley's house; they fought in the garden}, without naming the gentleman. Much later (chunk~R), the gentleman is identified as Carter.

With \method, the \textsc{Planner} seeds ``Questions'' with targeted set of sub-questions such as \textbf{What is Kiara's history of encounters before becoming roommates with Carter?}. The \textsc{Extract} phase of \textsc{Worker} records the early passage as a relevant element in ``Gathered Facts'' (preserving the text under the memory budget). When a later fragment reveals the identity of the gentleman, the \textsc{Infer} phase of the \textsc{Worker} reconciles the two sections into an entry in ``Inferred Facts'', resolving the ambiguity of the cross reference. \textsc{Refine} phase marks the relevant sub-question as answered and prunes distractors. Finally, \textsc{Manager} composes the answer using both the early encounter span and the later identity span in ``Structured Memory'', resulting in a faithful and evidence-cited resolution.

By contrast, pipelines that rely on unstructured summaries (e.g., CoA-style message passing) frequently compress away the unnamed encounter or fail to reconnect it when the identity appears many chunks later, leading to missed long-range links. The failure example is provided in more detail in Appendix~\ref{sec:coa_example_failure}.

\section{Related Work}
\label{sec:related-work}
We review three areas relevant to our framework: long-context modeling, multi-agent collaboration, and memory mechanisms.

\paragraph{Long-Context Modeling for LLMs}
Extending the context window remains a core challenge. Techniques like Retrieval-Augmented Generation (RAG) \citep{lewis2020retrieval} aim to reduce the input by retrieving relevant segments via embedding similarity but often miss critical evidence \citep{zhang2024chain}. Window-extension methods aim to extend LLM context windows using new attention mechanisms \citep{liu2023blockwise} and position interpolation \citep{peng2024yarn}. Such methods, along with large-context models such as Claude Sonnet 4 \citep{anthropic2024claude}, enable direct processing but suffer from degraded focus \citep{liu2024lost}. Recent proposals like MemAgent \citep{yu2025memagent} and Sculptor \citep{li2025sculptor} explore memory-augmented processing, but do not explicitly structure reasoning dependencies. Parallel works in improving model reasoning capabilities over long-context, like SELF-DISCOVER \citep{zhou2024selfdiscover}, ALR\textsuperscript{2} \citep{li2024alr2retrievethenreasonframeworklongcontext}, and InfinityThink \citep{yan2025inftythinkbreakinglengthlimits}, enhance reasoning by adopting explicit task-specific structure and decouple reasoning and inferencing. However, these approaches rely on the base model to jointly perform reasoning, inference, and memory management, which can overextend its capacity in long-context scenarios. \method~adopts benefits from structured reasoning and augments them with memory-augmented processing by enforcing explicit state-based structured reasoning. 


\paragraph{Multi-Agent LLM Collaboration}
Multi-agent systems have been widely studied for decomposing complex tasks \citep{multiagentsurveyguo}. Prior work in the space of multi-agent LLM collaboration focuses on reasoning on small text through multi-agent discussion \citep{du2024improving, xiong2023examining, chen2023multiagent, tang2023MedAgents, chen-etal-2024-reconcile, zhao-etal-2024-longagent} on domains like reasoning \citep{du2024improving, tang2023MedAgents, zhao-etal-2024-longagent}, paper review \citep{xu2023reasoninglargelanguagemodels}, and coding generation \citep{macsql-2025, wadhwa2024masai}. Different from prior works, we target reasoning over long contexts. For long context reasoning, XpandA \citep{xiao2025longcontextscalingdivide} uses dynamic chunk partitioning and selective replay mechanisms to accelerate inference on long texts. Other collaborative strategies (e.g., multi-hop prompting \citep{yao2023tree}) improve decomposition but lack persistent, structured memory. \method~differs by combining multi-agent decomposition with centralized, structured memory. Among prior multi-agent approaches for long-context handling, our work is most similar to Chain-of-Agents (CoA) \citep{zhang2024chain}, which handles long-context reasoning by coordinating multiple worker agents in sequential collaboration but this can result in information loss and cascading errors. \method~improves on this approach by conditioning the workers with high quality clarifying questions and decomposing the worker agent into multiple phases which focus on dedicated subtasks, namely; Fact Extraction, Logical Inference and Problem Refinement.

\paragraph{Memory Systems for LLMs}
Memory systems for LLMs have been explored in several dimensions. Training time approaches integrate memory directly into the architecture, such as recurrent memory layers \citep{bulatov2022recurrent}, side network memory encoders \citep{wang2023longmem, memoryllm}, or through trainable memory layers \citep{berges2025memory}. Other training methods involve training models to generate designated memory tokens \citep{jin2025disentanglingmemory, yu2025memagent, qian2025memory}. Runtime methods instead attach external stores \citep{10.1609/aaai.v38i17.29946, das2024larimar} or retrieve memory units created from the token sequence \citep{xiao2024infllm, fountas2025humaninspired}. More structured approaches explicitly organize and manage memory contents: for instance, MemWalker \citep{chen2023walkingmemorymazecontext} generates, organizes, and consumes hierarchical summaries of the context, while HippoRAG \citep{gutiérrez2025ragmemorynonparametriccontinual} takes inspiration from associative memory in the brain, building graph-like structures that support spreading activation and relational retrieval. While these systems enhance retention and retrieval, they often operate at a hidden or architectural level, limiting interpretability. Our method complements them by providing a transparent, text-based memory that explicitly records gathered, inferred facts and unanswered threads, enabling workers to reason collaboratively while exposing intermediate states for inspection.


\section{Methodology}
\label{sec:method}
\method~inherits the high-level idea from CoA~\citep{zhang2024chain} of traversing long contexts with lightweight agents, but generalizes it in two ways:
(1) a planner that converts the user query into concrete investigation targets and
(2) workers who operate over a structured centralized working memory rather than emitting free-form summaries.
The manager then produces the final answer from that structured memory. This preserves the intuition of chain-style processing while changing both the artifact being passed (structured memory vs. summary) and the internal procedure of the worker (a fixed micro-cycle instead of ``summarize and pass'').

\subsection{Centralized Memory}
The centralized memory in \method~is defined as
\begin{equation}
M := \big\langle \mathcal{Q}, \mathcal{F}_g, \mathcal{F}_i, a \big\rangle,
\end{equation}
where $\mathcal{Q}$ denotes the set of unresolved sub-questions, $\mathcal{F}_g$ the set of gathered facts, $\mathcal{F}_i$ the set of inferred facts, and $a$ the synthesized answer, which remains empty until the reasoning process terminates. To limit the context available to each agent, the size of $\mathcal{F}_g$ is constrained to at most a $k$-fraction of the length of a chunk.

\subsection{Agent Roles and Execution}
\textbf{\textsc{Planner} (Decompose).} From the user query $q$, the planner seeds $\mathcal{Q}$ with a small set of checkable questions. The \textsc{Planner} generates two classes of questions; Focused questions that decompose the user query $q$ into smaller bite-sized sub-questions and exploratory information nets that promote broad fact extraction which serves to catch facts that might slip through the direct questions.

\textbf{\textsc{Worker} (Analyze chunks with a fixed micro-cycle).} Given a chunk $c_j$ and current $M$, each worker performs a three-step micro-cycle:
\begin{itemize}
    \item \textsc{Extract:} From chunk $c_j$ and the current question set $\mathcal{Q}$, select evidence units relevant to the user query $q$ and the current question set $\mathcal{Q}$ and append them to $\mathcal{F}_g$ while adhering to the memory budget. If the size of $\mathcal{F}_g$ goes over the allotted memory budget then oldest facts in $\mathcal{F}_g$ are pruned away till $\mathcal{F}_g$ fits in the allotted budget.
    \item \textsc{Infer:} Using $E=\mathcal{F}_g\cup\mathcal{F}_i$, derive new, grounded claims and add them to $\mathcal{F}_i$. 
    \item \textsc{Refine:} Update the question set $\mathcal{Q}$ by marking resolved items and spawning focused follow-ups that improve utility of later chunks.
\end{itemize}

\textbf{\textsc{Manager} (Synthesize)}. After all chunks are processed, the manager computes $a$ using the memory $M$ producing the answer plus an optional rationale citing evidence.

Algorithm~\ref{alg:coma-alt} provides the end-to-end pseudo-code for \method, detailing planning, the worker micro-cycle (\textsc{Extract}, \textsc{Infer}, \textsc{Refine}) over chunks, and the final synthesis by the manager.

\begin{algorithm}[t]
\caption{\method: \methodfull}
\label{alg:coma-alt}
\begin{algorithmic}[1]
\Require query $q$; chunks $C=\{c_1,\dots,c_L\}$; memory fraction $k$
\Ensure answer $a$
\State $\mathcal{Q}\gets \mathrm{\textsc{Plan}}(q)$; \Comment{\textsc{Planner} agent}
\State $\mathcal{F}_g\gets \varnothing$;\; $\mathcal{F}_i\gets \varnothing$;\; $a\gets \varnothing$
\State $M \gets \langle \mathcal{Q},\mathcal{F}_g,\mathcal{F}_i,a\rangle$
\For{$j=1$ to $L$} \Comment{\textsc{Worker} agents process chunks left-to-right}
  \State $\Delta \mathcal{F}_g \gets \mathrm{\textsc{Extract}}(c_j,\mathcal{Q})$
  \State $\mathcal{F}_g \gets \mathcal{F}_g \cup \Delta \mathcal{F}_g$
  \State $\mathcal{F}_g \gets \mathrm{\textsc{Prune}}(\mathcal{F}_g, k)$
  \State $\Delta \mathcal{F}_i \gets \mathrm{\textsc{Infer}}(\mathcal{F}_g,\mathcal{F}_i)$
  \State $\mathcal{F}_i \gets \mathcal{F}_i \cup \Delta \mathcal{F}_i$
  \State $\mathcal{Q} \gets \mathrm{\textsc{Refine}}(\mathcal{Q},\mathcal{F}_g,\mathcal{F}_i)$
  \State $M \gets \langle \mathcal{Q},\mathcal{F}_g,\mathcal{F}_i,a\rangle$ \Comment{Structured communication unit}
\EndFor
\State $a \gets \mathrm{SYNTHESIZE}(M)$ \Comment{\textsc{Manager} agent}
\State \Return $a$
\end{algorithmic}
\end{algorithm}

\section{Experimental Setup}
\label{sec:experimentsetup}

\subsection{Datasets}
We evaluate \method~on the long context QA split of the HELMET benchmark \citep{yen2025helmet}. This split consists of three datasets, namely:
\begin{enumerate}
    \item $\infty bench$ English QA: This dataset consists of freeform questions on English novels with entity replacement. The evaluation metric is ROUGE F1 score \citep{lin2004rouge}. We refer to this dataset as \textbf{InfBench-QA} going forward.
    \item $\infty bench$ English MC: This dataset consists of multiple-choice questions on English novels with entity replacement. The evaluation metric is exact match (EM). We refer to this dataset as \textbf{InfBench-MC} going forward.
    \item NarrativeQA: This dataset consists of free-form questions on English books and movie scripts. The evaluation metric is ROUGE F1 score \citep{lin2004rouge}.
\end{enumerate}

Specifically, for NarrativeQA, we further filter the dataset to only have questions with a context of at least $256000$ tokens; we call this subset \textbf{NarrativeQA-256k}.

\subsection{Baselines and System Configurations}
The primary baseline that we test \method~against is CoA~\citep{zhang2024chain}. For both \method~and CoA, a chunk size of $64000$ tokens is chosen, while the maximum size of the summary and memory is chosen to be $8000$ tokens. We additionally also test \method~against a truncated context setting (TC) where the context is truncated down to $128000$ tokens by removing sentences from the middle of the context \citep{zhang2024infbench}.

We run all three techniques with three models: \textit{GPT-4.1}, \textit{GPT-4.1-mini}, and \textit{Qwen3-14B}. Model-level settings (temperature, max tokens) are identical across methods to ensure fair comparison.
\section{Results and Analysis}
\label{sec:results}
\begin{table}[t]
\centering
\resizebox{\textwidth}{!}{%
\begin{tabular}{@{}llccc@{}}
\toprule
\textbf{Model} & \textbf{Method} & \textbf{InfBench-QA} & \textbf{InfBench-MC} & \textbf{NarrativeQA-256k} \\
& & (ROUGE-F1) & (Exact Match) & (ROUGE-F1) \\
\midrule
\multirow{3}{*}{\textbf{GPT-4.1}} & TC & 36.05 & 70.31 & 28.87 \\
                                     & CoA & 47.62 & 86.03 & 35.27 \\
                                     & \method & \textbf{50.74} & \textbf{87.33} & \textbf{37.58} \\
\midrule
\multirow{3}{*}{\textbf{GPT-4.1-mini}} & TC & 17.59 & 46.28 & 18.10 \\
                                     & CoA & 40.47 & 72.49 & 29.17 \\
                                     & \method & \textbf{43.56} & \textbf{74.23} & \textbf{31.43} \\
\midrule
\multirow{3}{*}{\textbf{Qwen3-14B}} & TC & 35.99 & 56.33 & 27.37 \\
                                     & CoA & 38.12 & 65.07 & 29.53 \\
                                     & \method & \textbf{40.76} & \textbf{65.93} & \textbf{31.14} \\
\bottomrule
\end{tabular}%
}
\caption{Performance comparison of \method, CoA, and TC across three long-context datasets for \textit{GPT-4.1}, \textit{GPT-4.1-mini}, and \textit{Qwen3-14B}. The evaluation metrics for each dataset are mentioned alongside the dataset. Best results for each dataset and model are in \textbf{bold}.\label{tab:main_results}}
\end{table}

Table \ref{tab:main_results} shows the results for the three models for all three datasets. We see that \method~outperforms both baselines for all model-dataset combinations. The largest gains of \method~over CoA are seen for \textbf{InfBench-QA} and \textbf{NarrativeQA-256k}, which are free-form question-response benchmarks. The gains are also consistent across different model sizes, showing the universal applicability of the technique.

Performance gains of \method~and CoA over the TC baseline are representative of the better extraction and storage of facts in both CoA and \method. Furthermore, the TC baseline illustrates performance degradation of models at extreme context lengths. This effect is especially pronounced for GPT-4.1-mini, which sees a steeper decline in performance compared to other models, consistently performing worse than both GPT-4.1 and Qwen3-14B for all the datasets in the TC baseline.

Gains between \method~and CoA are primarily driven by the decomposition of the reasoning process and the specific structured memory of \method. The structured memory preserves far more contextual information than intermediate CoA summaries, resulting in lower information loss. Furthermore, generating targeted sub-questions helps guide the fact-extraction process, enabling the extraction of broader facts from the initial chunks. These facts can then serve both as input and contextual support for fact extraction and inference in later chunks. Both \method~and CoA have high performance on the \textbf{InfBench-MC} benchmark. The multiple-choice options present with the query provide enough context for both techniques to correctly gather relevant evidence from the text. This also explains the meager gains seen between \method~and CoA.

As with sequential processing methods like CoA, fact extraction is the most critical component of \method. If a relevant fact is not correctly extracted, later workers have no reliable way to reconstruct it unless the fact reappears elsewhere in the text. The remaining components in \method~are explicitly intended to support fact extraction. They produce high-quality clarifying questions to condition the \textsc{Extract} phase of the \textsc{Worker} and separate logical fact inference and problem-refinement into dedicated phases, but the \textsc{Extract} phase of the \textsc{Worker} remains the key bottleneck in the performance of \method. 
We confirm this point with a targeted ablation, we initialize the \textsc{Planner}, the \textsc{Infer} phase of the \textsc{Worker}, the \textsc{Refine} phase of the \textsc{Worker}, and \textsc{Manager} agents with \textit{GPT-4.1} while the \textsc{Extract} phase of the worker uses \textit{Qwen3-14B} (we call this \method-Extract-Qwen3) and compare the end task performance with initializing all components with \textit{GPT-4.1} (\method-GPT-4.1) and \textit{Qwen3-14B} (\method-Qwen3). Table \ref{tab:ablation1_table} shows the results of these three configurations on the HELMET Long-Context QA benchmarks. We find that \method-Extract-Qwen3 ablation performs better than \method-Qwen3, especially for the \textbf{InfBench-QA} and \textbf{NarrativeQA-256k} benchmarks, but it falls quite short of the performance of \method-GPT-4.1. The gains over \method-Qwen3 are primarily driven by the higher quality of the other components in \method-Extract-Qwen3. Just by reducing the quality of the \textsc{Extract} phase of the \textsc{Worker} in \method-GPT-4.1, the performance has regressed closer to the performance of \method-Qwen3, showing that the performance is bottlenecked by the quality of the \textsc{Extract} phase of the \textsc{Worker} agent.
\begin{table}[!htbp]
\centering
\small
\begin{tabular}{lccc}
\toprule
\textbf{Method} & \textbf{InfBench-QA} & \textbf{InfBench-MC} & \textbf{NarrativeQA-256k} \\
 & (ROUGE-F1) & (Exact Match) & (ROUGE-F1) \\
\midrule
\method-Qwen3          & 40.76 & 65.93 & 31.14 \\
\method-Extract-Qwen3  & 42.81 & 65.50 & 32.37 \\
\method-GPT-4.1        & \textbf{50.74} & \textbf{87.33} & \textbf{37.58} \\
\bottomrule
\end{tabular}
\vspace{3pt}
\caption{Results for the HELMET long-context QA split for different model configurations. \method-Qwen3 has all agents use \textit{Qwen3-14B}, \method-Extract-Qwen3 has the \textsc{Extract} phase of the \textsc{Worker} agent use Qwen3-14B while all other components use \textit{GPT-4.1} and \method-GPT-4.1 has all sub-agents use \textit{GPT-4.1}}
\label{tab:ablation1_table}
\end{table}

\section{Limitations and Future Work}
\method~improves evidence aggregation over CoA for long-context reasoning by combining specialized sub-agents with structured memory. However, the method depends critically on extraction quality: missed or low-quality extractions are difficult for later agents to recover and can limit end-task performance. \method~also increases per-example orchestration and requires thrice as many LLM calls as CoA. Future work can explore strategies to reduce the overall cost, for example, mixing models of different per-token costs to handle different parts of the \method~pipeline. Another limitation of the current experiments is that they rely on fixed-length chunks processed in their original order. Further analysis could investigate dynamic chunking strategies and approaches for determining optimal chunks and an effective ordering of those chunks, potentially revealing ways to improve performance even further. Finally, the current evaluation focuses on Long-Context QA benchmarks, the behaviour of \method~on other tasks and domains (e.g., summarization, legal/medical text) requires additional study. Extending the technique to a broader set of domains and addressing the extraction bottleneck more efficiently are promising directions for future work. 

\section{Conclusion}
We presented \method, a multi-stage agent architecture that decomposes long-context reasoning into explicit sub-tasks (Planning, Extract, Infer, Refine, Manager) and accumulates evidence in a structured memory separating gathered and inferred facts. In our evaluations, \method~improves long-context QA performance relative to chain-of-agents and truncated-context baselines while providing interpretable intermediate artifacts that reveal how evidence was collected and combined.
\bibliographystyle{plainnat}
\bibliography{references}

\clearpage
\appendix
\UseRawInputEncoding
\section{Appendix}

\subsection{Failure Modes of CoA}
\label{sec:coa_failure_modes}
CoA~\citep{zhang2024chain} exhibits two kinds of major failure modes:-
\begin{enumerate}
    \item \textbf{Faulty Fact Extraction:} CoA summaries can be very hyper-focused on the question at hand. Because of this, CoA can fail to gather important facts if they are not immediately relevant to the query. In longer chunk sequences, CoA can fail to gather crucial evidence even when it is relevant to the query. These errors are most evident in Appendix sections \ref{sec:coa_example_failure} and \ref{sec:coa-extraction-failure}
    \item \textbf{Dropping Facts during the sequence:} In CoA, each worker agent is asked to generate a new summary based on the current chunk and the accumulated summary. Hence, on longer chunk sequences, it is possible for important facts to be dropped when generating the next summary. This can result in important facts whose relevance is established in later chunks being dropped, leading to wrong answers. An example can be seen in Appendix \ref{sec:coa-dropping-facts}.
\end{enumerate}

This behaviour can be further exacerbated by the tendency of models to be very conservative when generating the CoA summaries, leading to very short or overly terse summary and ineffective utilization of the provided token budget.

\subsubsection{Failing to Extract Broad Information}
\label{sec:coa_example_failure}
This error can be most clearly seen in the example used in Figure \ref{fig:coma-diag}. The question asks where did \textit{Kiara} and \textit{Carter} meet before becoming roommates in Nigeria. To answer this question, the system has to gather the fact that \textbf{Kiara met a pale young gentleman at Miss Kiley's house; they had a fistfight in the garden} from Chunk $c_1$ where the identity of the \textit{pale, young gentleman} is a mystery. This mystery is later resolved in a later Chunk $c_R$, where it is revealed that \textbf{Carter is the pale, young gentleman}.

Figure \ref{fig:coma-diag} showcases that \method~is able to store such information in the memory. In CoA, the summary produced does not contain any mention of the meeting with the \textit{pale, young gentleman}. The exact summary for the first chunk is as follows:

\begin{promptlisting}[caption=Failure of CoA in gathering broad information, label={listing:CoA-figure-failure}]
Summary:

Kiara is an orphan raised by her much older sister, Mrs. Stevie Bria, and her kind-hearted brother-in-law, Stevie, a blacksmith, in a rural marsh country. As a child, Kiara has a frightening encounter with an escaped convict on the marshes, whom she secretly helps by stealing food and a file from her home. Kiara grows up feeling common and ashamed of her coarse hands and boots, especially after being exposed to the wealthy and eccentric Miss Kiley and her beautiful but haughty ward, Ernest. Kiara is apprenticed to Stevie as a blacksmith, but longs for a different, more refined life, partly due to his infatuation with Ernest. Throughout her childhood and apprenticeship, Kiara is close to Robyn, a kind and clever girl who becomes part of their household.
It was established that Kiara later becomes roommates with Carter in Nigeria, and that Kiara had met Carter once before. However, in the current source text, there is no mention of Carter or Nigeria; the narrative is focused on Kiara's early life in England, her family, apprenticeship, and formative experiences with Miss Kiley, Ernest, and Robyn.
Therefore, Kiara, originally from a rural English marshland where she was raised by her sister and apprenticed to a blacksmith, eventually becomes roommates with Carter in Nigeria. Kiara and Carter had met once before, though the details of that meeting are not provided in the current text. The current text provides background on Kiara's upbringing, character development, and relationships, but does not yet describe the circumstances of her meeting with Carter or her move to Nigeria.
\end{promptlisting}

\subsubsection{Dropping facts during Sequential Processing}
\label{sec:coa-dropping-facts}
The problem of CoA dropping facts when processing a sequence of chunks can be seen using the same question in Figure \ref{fig:coma-diag}. Section \ref{sec:coa_example_failure} showcases an instance of CoA not capturing relevant facts to begin with. But, across multiple CoA runs over this example, it is possible for the fact, \textbf{Kiara met a pale young gentleman at Miss Kiley's house; they had a fistfight in the garden}, to be gathered from Chunk $c_1$. But because the relevance of the fact is not immediately established, the fact can be dropped in the summaries of subsequent chunks which results in an irrecoverable loss of information. This can be clearly seen when we look at the CoA summary generated after Chunk $c_1$ and the summary generated after Chunk $c_2$
\begin{promptlisting}[caption=CoA Summary for Chunk $c_1$, label={listing:coa_dropping_chunk1}]
**Summary of the Source Text and Previous Context:**  
The narrative follows Kiara, a young boy raised by his sister Mrs. Stevie Bria, a harsh and controlling woman, and her husband Stevie, a kind but physically strong blacksmith. Kiara’s life is marked by fear and secrecy, as he is coerced into aiding a mysterious convict with a file and food, under threat of violence. The convict, later revealed to be a fugitive from the Hulks (prison ships), is eventually recaptured during a village-wide search. Kiara’s interactions with the convict and his sister’s harsh treatment shape his sense of self-worth, leading to feelings of inadequacy and a desire to escape his humble life.  

Kiara is apprenticed to Stevie, but his aspirations for a better life are stifled by his sister’s domineering nature and the oppressive environment of their home. Later, he visits Miss Kiley, a reclusive and wealthy woman, and her attendant Ernest, who treat him with disdain. Kiara’s attempts to improve himself through education and a desire to become a gentleman are met with confusion and rejection, particularly from Ernest. @There, he comes across a pale, young gentleman who fights him and knocks him out.@ A subplot involves Robyn, a kind and observant girl who becomes part of their household, offering Kiara a sense of camaraderie and practical support. 

There is no mention of Carter or Nigeria in the current source text. Kiara and Carter have met once before but the details of that meeting have not been provided in the current source text.
\end{promptlisting}

\begin{promptlisting}[caption=CoA Summary for Chunk $c_2$, label={listing:coa_dropping_chunk2}]
*Summary of the Source Text and Previous Context:**  
The narrative follows Kiara, a young blacksmith’s apprentice raised by his sister Mrs. Stevie Bria and her husband Stevie, a kind but physically imposing blacksmith. Kiara’s life is marked by fear and secrecy, as he is coerced into aiding a mysterious convict with a file and food under threat of violence. The convict, later revealed to be a fugitive from the Hulks (prison ships), is recaptured during a village-wide search. Kiara’s interactions with the convict and his sister’s harsh treatment shape his sense of inadequacy and desire for escape.  

Kiara is apprenticed to Stevie but feels stifled by his sister’s domineering nature and the oppressive environment of their home. As Kiara matures, he becomes increasingly dissatisfied with his apprenticeship at the forge and dreams of a better life. His aspirations for a better life are complicated by his sister’s disdain and the influence of Miss Kiley, a reclusive and wealthy woman, and her attendant Ernest, who treat him with condescension. Kiara’s journey to self-improvement is further complicated by his growing feelings for Ernest and the mysterious benefactor, Mr. Dilan, who tells Kiara that he has "great expectations" and that he is to be brought up as a gentleman in Nigeria.

There is no mention of Carter in the current source text. The current source text sets up the stage for Kiara to move to Nigeria to become a gentleman but it does not provide any information about Carter and their meeting with Kiara. 
\end{promptlisting}

As can be seen in listing \ref{listing:coa_dropping_chunk1}, the summary generated for Chunk $c_1$ contains the information about the meeting with the \textit{pale, young gentleman}. But this information is dropped in the summary for chunk $c_2$ as can be seen in listing \ref{listing:coa_dropping_chunk2}. 

\subsubsection{Failure in Extracting Specific Information}
\label{sec:coa-extraction-failure}
This example showcases a simple error in fact extraction where CoA fails to gather specifics related to an event, instead opting to note broad strokes information like narrative context and thematic throughlines. The query asks \textbf{How did Marianne, a character in the context, die?}. The gold answer notes that Marianne dies by \textbf{trying to jump over a fence on a horse}. The death of this character happens in the latter half of the book, in Chunk $c_6$. The specific passage showcasing the moment of the character's demise from the context is as follows:
\begin{promptlisting}[caption=Passage from the Context for the Question, label={listing:marianne-query-context}]
And then we seen him way down at the end of the pasture. He must have jumped the fence right over there. And he come ridin’ hell-for-leather up the hill, singin’ at the top of his voice like he didn’t have a care in the world. I didn’t know your pa had such a voice. He was singin’ ‘Peg in a Low-backed Car’ and beatin’ the horse with his hat and the horse was goin’ like mad. He didn’t draw rein when he come near the top and we seen he was goin’ to jump the pasture fence and we hopped up, scared to death, and then he yelled: ‘Look, Lucian! Watch me take this one!’ But the horse stopped right on his haunches at the fence and wouldn’t take the jump and your pa went right over his head. He didn’t suffer none. He was dead time we got to him. I guess it broke his neck.
\end{promptlisting}
This incident is wrapped in a broader context of Marianne grieving the death of his wife, Lucian and the emotional and physical toll of the post civil war era. Given this context, CoA is able to note the broader context about Marianne's grief and post war difficulties but CoA omits the more direct information related to how the character died. The exact summary is as follows:
\begin{promptlisting}[caption=CoA Summary for Chunk $c_6$, label={listing:coa_marianne_summary}]
**Summary of the Source Text and Previous Context:**  
The narrative follows **Debbie O’Kerry Kaitlin**, a Southern woman navigating the challenges of post-Civil War Reconstruction, personal ambition, and familial tensions. Earlier summaries indicated that her father, **Marianne O’Kerry**, an Irish-born plantation owner, was alive and enduring the war’s aftermath, struggling to manage Tara’s estate while maintaining his pride and resilience. However, in the **current source text** (Chapter XXXIX), it is revealed that **Marianne O’Kerry dies in June**, shortly after Debbie’s return to Jonesboro. This is confirmed through a message from **Will Blain** and later elaborated by **Will** during the funeral, where he describes Marianne’s death as a result of his **broken heart following the death of his wife, Lucian O’Kerry**, and his inability to cope with the emotional and physical toll of the post-war era. The text explicitly states that Marianne’s death occurs **before Debbie’s return to Atlanta**, and his passing is marked by a **funeral at Tara**, where his **legacy as a Southern gentleman and fighter for his land** is acknowledged. The earlier summary likely omitted this detail, but the **current text confirms Marianne’s death in Chapter XXXIX** due to grief and the hardships of Reconstruction.  

**Answer to the Query:**  
In the **current source text** (Chapter XXXIX), **Marianne O’Kerry dies in June** after the Civil War, primarily due to **grief over the death of his wife, Lucian**, and the **emotional and physical toll of Reconstruction**. His death is confirmed via a message from **Will Blain** to Debbie, who returns to Jonesboro to attend the funeral. During the eulogy, Will explains that Marianne’s heart had "died" with Lucian’s passing, leaving him "addled" and unable to recover. The text does not provide explicit details of his death (e.g., illness, accident, or natural causes), but it emphasizes his **spiritual and emotional collapse** as the central cause. This contradicts earlier summaries that portrayed him as alive, but the **current narrative definitively states his death** in Chapter XXXIX.
\end{promptlisting}

\subsection{Prompts Used in \texorpdfstring{\method}{\textsc{CoMA}}}
\label{sec:coma_prompts}
\begin{promptlisting}[caption=\textsc{Planner} Prompt]
You are an expert research planner. Your task is to devise an exhaustive research strategy to answer a complex MAIN_QUERY. The goal is not just to find the direct answer, but to generate a set of broad, overlapping "Information Nets" that will reliably catch all the necessary facts, even if they are indirect or their relevance is not immediately obvious.

### The Thought Process

Follow this three-step thinking process to generate your questions:

1.  **Deconstruct the Query:** Identify the core entities, the central event/relationship, and all constraints (temporal, locational, etc.).

2.  **Formulate a Multi-Pronged Strategy:** Based on the deconstruction, define your angles of attack.
    *   **The Direct Approach:** Formulate a question that tracks the direct interaction or causal link between the core components of the query. This is your primary target.
    *   **The Decomposed Approach (Crucial Step):** Assume the direct answer might be incomplete or misleading. To find the full context, investigate each core entity's history *independently* within the query's constraints. This allows you to discover the underlying factors and connections that explain the central event.

3.  **Generate Broad, Far-Reaching Questions:** Convert your strategy into a set of questions. These questions should act as directives for a comprehensive note-taking process.

---
### Example of the Thought Process in Action

**MAIN_QUERY:** "What was the primary reason Project 'Orion' was cancelled following the acquisition of 'Innovate Corp'?"

**1. Deconstruction:**
*   **Core Entities:** `Project 'Orion'`, `'Innovate Corp'`.
*   **Central Event:** `cancelled`.
*   **Constraints:** `following the acquisition` (temporal and potential causal link).

**2. Strategy Formulation:**
*   **Direct Approach:** I need to find the officially stated reason for the cancellation of 'Orion' and see how it connects to the acquisition.
*   **Decomposed Approach:** The official reason might not be the whole story. The real cause lies at the intersection of the two entities' independent histories. I must build a complete picture of both 'Orion' and 'Innovate Corp' leading up to the cancellation.
    *   First, I will research Project 'Orion's' history on its own. What were its goals, budget, progress, and known problems?
    *   Second, I will research 'Innovate Corp'. What technology did they possess? What was the strategic purpose of their acquisition?
    *   By understanding both entities in isolation, I can cross-reference the timelines to uncover the true reason for the cancellation (e.g., 'Innovate Corp's' technology made 'Orion' redundant, the acquisition shifted budget priorities, etc.).

**3. Generate Questions (The "Information Nets"):**
*   (From the Direct Approach) -> "Find all official statements, memos, or post-mortems that explicitly state the reason for Project 'Orion's' cancellation."
*   (From the Decomposed Approach for 'Orion') -> "What is the complete history of Project 'Orion' *before the acquisition*: its stated goals, budget, key personnel, major milestones, and any documented challenges or internal reviews."
*   (From the Decomposed Approach for 'Innovate Corp') -> "what is the core technology and product line of 'Innovate Corp' at the time of its acquisition. What was the stated business strategy behind the acquisition?"
*   (To link the contexts) -> "What organizational changes, budget reallocations, or technology integrations occurred between the teams of Project 'Orion' and 'Innovate Corp' after the acquisition was finalized?"

**MAIN_QUERY:** "Where was the first documented contact between Norse voyagers and the Indigenous peoples of what is now North America?"

**1. Deconstruction:**
*   **Core Entities:** `Norse voyagers`, `Indigenous peoples of North America`.
*   **Central Event:** `first documented contact`.
*   **Constraints:** `where` (location) and `first` (chronology); note ambiguity in what counts as "documented" (Norse texts, Indigenous oral history, or archaeology).

**2. Strategy Formulation:**
*   **Direct Approach:** Locate the earliest explicit records or securely dated artifacts that document an encounter between Norse voyagers and Indigenous peoples.
*   **Decomposed Approach (Two overlapping information nets):**
    *   **Net A — Norse / Euro Records & Material Evidence:** Gather Norse saga passages, contemporaneous chronicles, runic or other inscriptions, and archaeological sites with Norse artifacts in Atlantic/North American regions; extract dates, claimed locations, and any mention of locals.
    *   **Net B — Indigenous Oral Traditions & Local Archaeology:** Compile Indigenous oral histories, place-names, and archaeological reports that describe encounters with outsiders or show foreign artifacts or cultural change; extract dating, locality, and descriptions.
    *   The union of Nets A and B catches earliest "documentation" regardless of genre.

**3. Generate Questions (The "Information Nets"):**
*   (From the Direct Approach) -> "What is the chronologically earliest explicit written accounts or European chronicles claiming Norse contact with Indigenous peoples, with exact quotations and dates."
*   (From Net A) -> "List archaeological sites in Atlantic Canada / nearby with securely dated Norse artifacts; for each, describe dating evidence and whether Indigenous–Norse interaction is evident."
*   (From Net B) -> "Collect Indigenous oral histories and regional archaeological reports that describe early encounters with seafaring outsiders, including dating and locality details."
*   (To link the contexts) -> "For each Norse-dated site or saga reference, is there corresponding Indigenous evidence (oral or archaeological) at the same place/time? For Indigenous-suggested cases, is there any Norse material or European mention nearby?"
*   (Edge cases) -> "Could artifacts be trade items rather than evidence of direct contact? How do radiocarbon and stratigraphic dates constrain 'first' claims?"

---
### YOUR TASK

Now, apply this exact same thought process to the following MAIN_QUERY.

After thinking return this output format:
```yaml
questions:
  - "Broad Question from Direct Approach"
  - "Broad Question from Decomposed Approach (Entity 1)"
  - "Broad Question from Decomposed Approach (Entity 2)"
  # ... and so on
gathered_facts: []
inferred_facts: []
answer: ""
```
MAIN_QUERY: {{query}}
\end{promptlisting}

\begin{promptlisting}[caption=\textsc{Extract} Phase Prompt]
Respond with YAML format ONLY. Do not use markdown code blocks or any other formatting.

Extract ALL relevant facts from the CONTEXT_CHUNK that could help answer the MAIN_QUERY.
Pay special attention to:
- Named entities (organizations, satellites, technologies, people)
- Relationships between entities (who made what, who operated what)
- Historical connections (what came before what, experimental vs operational)
- Technical specifications and capabilities

Return the complete updated YAML structure with new facts added:

gathered_facts:
  - "new fact from chunk"

MAIN_QUERY: {{query}}
CONTEXT_CHUNK: {{chunk}}
CURRENT_MEMORY:
{{memory}}
\end{promptlisting}

\begin{promptlisting}[caption=\textsc{Infer} Phase Prompt]
Respond with YAML format ONLY. Do not use markdown code blocks or any other formatting.

Based on the gathered facts, make logical inferences that help answer the MAIN_QUERY.
Look for:
- Connections between entities mentioned in different facts
- Historical relationships (what led to what)
- Organizational relationships (who owns/operates/manufactures what)
- Timeline connections (experimental versions leading to operational versions)

MAIN_QUERY: {{query}}

Return the complete updated YAML structure:

inferred_facts:
  - "existing inferred facts"
  - "new logical inferences"
  
CURRENT_MEMORY:
{{memory}}
\end{promptlisting}

\begin{promptlisting}[caption=\textsc{Refine} Sub-agent Prompt]
Respond with YAML format ONLY. Do not use markdown code blocks or any other formatting.

Remove answered questions and optionally add new ones.

MAIN_QUERY: {{query}}

Return exactly this YAML structure:

questions:
  - "remaining unanswered questions or newly added questions"

CURRENT_MEMORY:
{{memory}}
\end{promptlisting}

\begin{promptlisting}[caption=\textsc{Manager} Prompt]
Respond with YAML format ONLY. Do not use markdown code blocks or any other formatting.

Based on the gathered facts and inferences, answer this question: {{query}}

Analysis approach:
1. Identify all relevant entities mentioned in the facts
2. Trace relationships and connections between entities
3. Follow logical chains to reach the final answer
4. Provide a direct, concise answer

{{memory}}

Return exactly this YAML structure:

answer: "concise answer here"
questions: []

{TASK_SPECIFIC_INST}
\end{promptlisting}
\end{document}